# Conditional Deep Learning for Energy-Efficient and Enhanced Pattern Recognition

Priyadarshini Panda, Abhronil Sengupta, and Kaushik Roy
School of Electrical and Computer Engineering, Purdue University
{pandap, asengup, kaushik}@purdue.edu

*Abstract*—Deep learning neural networks have emerged as one of the most powerful classification tools for vision related applications. However, the computational and energy requirements associated with such deep nets can be quite high, and hence their energy-efficient implementation is of great interest. Although traditionally the entire network is utilized for the recognition of all inputs, we observe that the classification difficulty varies widely across inputs in real-world datasets; only a small fraction of inputs require the full computational effort of a network, while a large majority can be classified correctly with very low effort. In this paper, we propose Conditional Deep Learning (CDL) where the convolutional layer features are used to identify the variability in the difficulty of input instances and conditionally activate the deeper layers of the network. We achieve this by cascading a linear network of output neurons for each convolutional layer and monitoring the output of the linear network to decide whether classification can be terminated at the current stage or not. The proposed methodology thus enables the network to dynamically adjust the computational effort depending upon the difficulty of the input data while maintaining competitive classification accuracy. We evaluate our approach on the MNIST dataset. Our experiments demonstrate that our proposed CDL yields 1.91x reduction in average number of operations per input, which translates to 1.84x improvement in energy. In addition, our results show an improvement in classification accuracy from 97.5% to 98.9% as compared to the original network.

## I. INTRODUCTION

For many computer vision applications, all inputs are not created equal. Consider the simple example of recognizing a person from two images: one where the person is standing against a plain blue backdrop and other where he is in the midst of a crowd. Clearly, the latter one takes more time and effort. Ideally, to obtain both speed and energy efficiency, computational time and energy used by algorithms should be proportionate to the difficulty of the input instances [1]. Unfortunately, for most applications, separating easy inputs from difficult ones at runtime is challenging. Thus, conventional algorithms and architectures expend equal effort on all inputs. In this work, we focus on a particular class of machine-learning algorithms, Deep Learning Convolutional Neural Networks (DLN) [2,3], and show how they can be used to construct a cascaded architecture for conditional activation of the latter layers in the network depending upon the difficulty of the input data, for faster and more energy-efficient implementations.

DLNs have proven to be very successful for many real-world applications such as Google Image search [4,5], Google Now speech recognition [6,7], and Apple Siri voice recognition [8] among others. However, being large scale and densely connected makes them highly computationally intensive. For instance, SuperVision [9], a DLN that won the ImageNet visual recognition challenge, demands compute performance in the order of 2-4 GOPS per classification [10]. This energy consumption can be considerably reduced if the DLNs can scale their computational effort based on the input data. Interestingly, we observe that the convolutional layers (CNN layers) of a DLN, interpreted as visual layers, learn a hierarchy of features which transition from general (similar to Gabor filters and color blobs [11]) to specific as we go deeper into the network [12]. In fact, DLN models that are trained for classification have been used as feature extractors by removal of the final output layer [13-15]. In particular, features extracted from a pre-trained DLN, OverFeat [16], have been successfully used in computer vision tasks such as scene recognition or object detection. Here, we intend to utilize the generic-to-specific transition in the learnt features of the CNN layers to identify the inherent variability in the difficulty of the inputs in a dataset. The outputs of the first layers of a DLN are used to classify the easy instances of a given dataset without activating the latter layers of the network. Only for the hard instances that in general, constitute a small fraction of the dataset, the deeper layers are enabled to make more accurate classifications. Thus, we exploit the usefulness of the CNN features to introduce Conditional Deep Learning (CDL) for energy-efficient pattern recognition with competitive classification accuracy as compared to the original DLN.

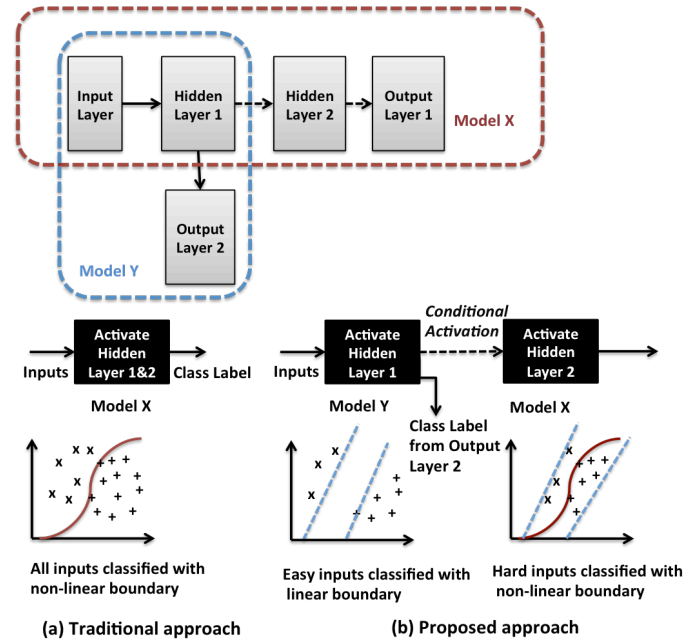

Fig. 1. (a) Traditional approach where both layers are activated and all inputs are classified with a non-linear boundary (b) Proposed approach where easy instances are classified at hidden layer 1 with linear boundary and hard instances at 2$^{nd}$ layer with non-linear boundary [1].

This work was supported in part by C-SPIN, one of the six centers of STARnet, a Semiconductor Research Corporation Program, sponsored by MARCO and DARPA, by the Semiconductor Research Corporation, the National Science Foundation, Intel Corporation and by the National Security Science and Engineering Faculty Fellowship.

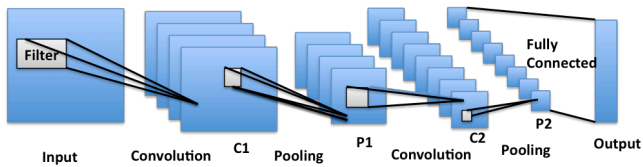

Fig. 2. A standard architecture of a Deep Learning Convolutional Network.

DLNs alike other supervised learning approaches have two modes of operation: training and testing. In the training phase, decision boundaries are constructed with training labels provided with the dataset. In the testing phase, the trained model is used to classify new instances. The basic methodology of CDL is as follows: During training, we construct a series of decision models (i.e. cascade of linear networks at every convolutional layer). This is completely different from the traditional approach where a single complex model (i.e. baseline DLN) is only used. In the test phase, the difficulty of the input instance determines the number of models or linear networks to be applied for accurate classification. Fig. 1 illustrates our methodology with a 2-hidden layer artificial neural network classifier. Fig. 1(a) shows the traditional approach where input training instances are classified into two categories by the complex model X. It is evident that non-linear boundaries would require more number of hidden layers and would thus be more computationally intensive than the linear boundary models. In the example of Fig. 1(a), the model X requires activation of both the hidden layers to classify the instances with high accuracy. However, this causes redundant activation of the second hidden layer for the easy instances of the dataset which increases the computational effort. We address this inadequacy with our proposed approach shown in Fig. 1(b). It consists of two decision models (Y and X) created by adding an output layer 2 after the 1$^{st}$ hidden layer. The simpler model Y (only 1$^{st}$ layer activated) selectively classifies only the easy training inputs that lie further from the original non-linear boundary by creating a hyperplane or region around it. If the confidence level of the output layer 2 for a given instance is below a certain threshold δ, the complex model X is employed by activating the 2$^{nd}$ hidden layer. Thus, the 2$^{nd}$ layer is only activated for the hard instances in the dataset. This approach thus leads to substantial energy savings, since the complex decision boundary (non-linear model X) need not process all data instances.

## II. CONDITIONAL DEEP LEARNING CLASSIFICATION

In this section, we present our structured approach to design the proposed Conditional Deep Learning Network (CDLN). As introduced earlier, CNNs form the basis of a deep learning network. DLN consists of one or more pairs of convolution and max pooling layers [17]. Fig. 2 shows a basic DLN structure with convolutional layers (C1, C2) followed by pooling layers (P1, P2). A convolution layer convolves a set of weight kernels with a portion of the previous layer to obtain an array of output maps. These kernels are repeated across the entire input space. A max-pooling layer lowers the dimensionality of the activations in the convolution layer by taking the maximum activation in a particular window of the previous layer map. This incorporates translational invariance in the DLN to small variations in positions of input images. Deeper layers require larger number of kernels that work on lower dimensional inputs to process complex components of the image. The final fully connected layers combine inputs from all maps in the preceding layer and perform overall classification of the input data. This hierarchical structure gives good results for image recognition tasks [18]. As mentioned earlier, CNN layers of DLN models that are trained for classification, have been used as feature extractors by removal of the the output layer. We exploit the efficacy of the convolutional layer features to develop an architecture in which easy instances can be classified earlier without activating the latter layers of the DLN network.

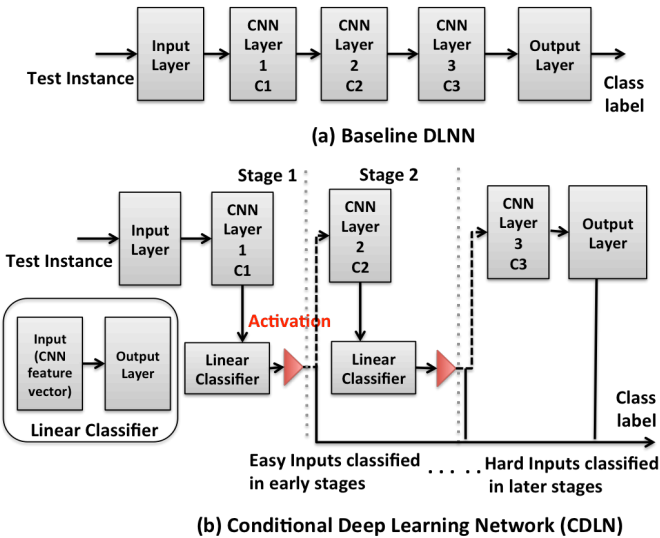

Fig. 3. (a) Baseline deep learning network (b) CDLN with linear classifiers added at the convolutional layers whose output is monitored to decide if classification can be terminated at current stage or not.

Fig. 3 shows the conceptual view of CDLN. Fig. 3(a) consists of the baseline deep learning network with 3 convolutional layers (CNN layers: C1, C2, C3) which are learnt using the standard backpropagation algorithm. We have not shown the pooling layers or the filters for the sake of convenience in representation. Fig. 3(b) illustrates our cascaded approach wherein the output features from each convolutional layer are fed to a linear classifier. The linear classifiers consist of same number of output neurons as that of the output layer of the baseline DLN (Fig. 3(a)). Thus, our proposed methodology consists of several stages, equivalent to the number of CNN layers, connected in a sequence. Each stage contains a linear classifier trained on the convolutional layer features corresponding to that stage. Depending upon the output of the linear classifiers in a given stage, the following stage of the CDLN is enabled. As mentioned earlier, as we go deeper into the network, the decision boundary model for each stage in the CDLN becomes progressively non-linear. Thus, class labels are produced at stage 1 for easy inputs and latter stages for hard ones.

Besides the linear classifiers, the stage consists of an activation module (triangles in Fig. 3(b)). During test time, input instance is passed through each stage to produce a class label. The stage also generates a confidence value along with the class label. The activation module utilizes this confidence to decide if the input instance should get classified in the current stage or passed to the next stage. This decision is based on the following two criteria:

- If the linear classifiers do not produce sufficient confidence associated with any of the class labels or produce a sufficient confidence for more than one label, the input is deemed to be difficult to classify by the current stage and is passed along to the next stage.

- If the linear classifiers produce sufficient confidence associated with only one label, then the classification process is terminated at that stage and the corresponding label is produced as output of the framework.

In addition to energy-efficiency, we also observe that the performance of the CDL network is better than the baseline DLN in terms of classification accuracy. This can be attributed to the fact that the linear networks being small scale with few neurons and synapses can be trained rapidly and easily to achieve better least mean square error as compared to the baseline DLN. Hence, a DLN, which is less than optimal, i.e. not fully trained or over-fitting, can also extract features for the linear networks that yield competitive classification accuracy.

## A. Efficiency and Accuracy Optimization

As the CDLN is composed of many individual stages, comprising of a series of linear classifiers, the following two factors determine their overall efficiency and accuracy: (a) the number of linear classifiers added and (b) the fraction of inputs processed at each stage. These factors present a fundamental tradeoff in the CDLN design methodology. The tradeoffs are discussed in the following subsections.

### A.1 Adding linear classifiers at the convolutional layers

First, we examine whether it is desirable to add a linear classifier for every convolutional layer of the DLN. Please note that we need to take into account the additional cost of adding an output layer of neurons for each convolutional layer while calculating energy costs [1]. Let the computational cost of the CDLN at a particular stage or layer $i$ be $\gamma_i$ per instance. Let the fraction of instances that reach stage $i$ be $I_i$. Similarly, the fraction of instances that reach stage $i+1$ is $I_{i+1}$. Thus, the stage $i$ classifies only a smaller subset $(I_i - I_{i+1})$ of the inputs. Stage $i$ should satisfy Eq. 1 shown below in order to improve the overall efficiency of the framework.

$$(\gamma_{i+1} - \gamma_i).(I_i - I_{i+1}) > \gamma_i.I_{i+1} \quad (1)$$

The left hand side of Eq. 1 signifies the efficiency improvement due to the addition of the linear classifier at the convolutional layer $i$, which is the product of the fraction of inputs classified at the stage $(I_i - I_{i+1})$ and the reduction in cost comapred with the activation of the convolutional layer in the next stage. The left side product should be larger than the right side of Eq. 1, which represents the penalty that addition of the linear classifier inflicts on instances that are misclassified i.e. if the linear classifier was not present then the instances $(I_{i+1})$ can be classified directly by the linear classifier corresponding to the convolutional layer in the next stage, $i+1$.

### A.2 Modulating activation of layers using confidence value

The activation module described earlier uses the confidence value of the output at each stage to selectively classify the input or pass it to the next stage. To better understand how the confidence influences the CDLN, consider the example shown in Fig. 4. Each stage in the Fig. 4 is defined as shown in Fig. 3(b) i.e. it consists of a convolutional layer (from the baseline DLN) which is fed as input to a linear classifier. The CDLN classifies a given input instance into one of the four class labels. During test time, data is processed through each stage of the CDL network to produce a class label. The linear classifiers in the CDLN, in addition to the class label, provide a measure of confidence (e.g. class probabilities or distance from the decision boundary).

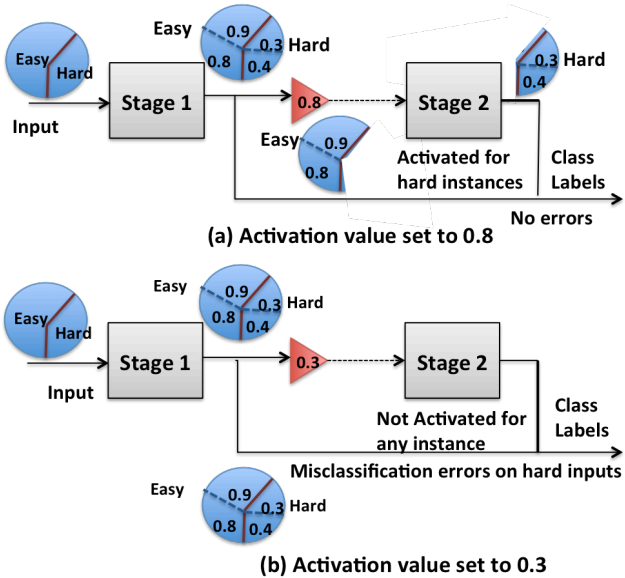

Fig. 4. (a) Activation value set to 0.8 terminates the classification for easy instances at stage 1 and enables stage 2 for hard instances (b) Activation set to 0.3 terminates the classification for easy and hard instances at stage 1.

Referring to Fig. 1 (b), the confidence value, thus, defines a region around the initial or original non-linear decision boundary that separates the easy and hard instances. Hence, a low confidence value output at any stage implies that the given test instance is a hard input and needs to be processed by the latter layers of the network for accurate classification. Fig. 4(a) shows that stage 1 gives a confidence level of 0.95 and 0.8 associated with a fraction of easy instances while 0.3 and 0.4 for the hard instances. Choosing the activation value of 0.8 would terminate the classification process at stage 1 for the easy instances and activate the latter stage only for the hard inputs. If the value is chosen to be 0.9, then the second stage will be redundantly activated for a fraction of easy instances (those with confidence value of 0.8) resulting in a decline in efficiency. On the other hand, if the activation value is set to 0.3 as in Fig 4(b), then, the second stage is not enabled at all. However, due to the low confidence value, majority of the hard instances will be misclassified. This would result in a significant decline in accuracy. Thus, modulating the activation value allows us to control the efficiency and accuracy of the framework. For computational efficiency, we choose higher confidence values (around 0.5 -0.7) during training to avoid misclassification errors.

## III. DESIGN METHODOLOGY

In this section, we describe the procedure for training and testing the CDLN.

### A. Training the CDLN

Algorithm 1 shows the pseudo code for training the CDLN. The process takes the original DLN $N_{orig}$, training data $D_{tr}$ with the corresponding labels as input and produces a conditional deep learning network $N_{cdl}$ with the optimized number of stages.

**Algorithm 1:** Methodology to train the CDLN
**Input**: Original DLN $N_{orig}$, training dataset $D_{tr}$ with the target labels
**Output**: CDLN $N_{cdl}$ including the optimum number of stages
1. Train $N_{orig}$ using $D_{tr}$ and obtain the classifier cost $\gamma_{orig}$
2. **initialize** count= # of Convolutional layers in $N_{orig}$, $G_i$= +infinity
3. **while** ($G_i > \varepsilon$)
4. **for** i=1:count **do**
5. Obtain the CNN features for the given $D_{tr}$ corresponding to all maps for each convolutional layer of the DLN.
6. Concatenate the CNN features into a 1-D vector and feed it as input to a linear classifier ($LC_i$) with the same number of output neurons as $N_{orig}$.
7. Train $LC_i$ with the target labels from $D_{tr}$ using the least mean square rule.
*//decide if a linear classifier needs to be added from second CNN layer or stage onwards*
8. **initialize** $I_i$= # of input instances that reach stage $i$, $Cl_i$= # of instances classified at stage $i$
9. $G_i=(\gamma_{orig} - \gamma_i).Cl_i - \gamma_i.(I_i - Cl_i)$
10. **if** $G_i > \varepsilon$ **then** admit $LC_i$ into $N_{cdl}$
11. **end for**
12. **end while**

The baseline DLN ($N_{orig}$) is first learnt for a given training set (step 1). The learnt CNN features corresponding to the training data are concatenated into a 1-dimensional vector and given as input to the linear classifier at every stage. The linear classifiers ($LC_i$) are then trained on the same training data using the least mean square rule (steps 4-7). For each layer/stage from the second layer onwards, we compute the gain $G_i$ which is the difference between the increase in efficiency for the instances classified at stage $i$ and the additional cost it inflicts on instances that are passed to the next stage (step 9). We add the linear classifier $LC_i$ to the CDLN $N_{cdl}$ if $G_i$ exceeds a certain user-defined threshold ε (step 10). The algorithm terminates if addition of an output layer at a particular CNN layer (i.e. a linear classifier) does not improve the overall gain $G_i$ beyond ε (step 3). Please note that the linear classifiers being small scale converge to the global minima (least error attainable by the linear classifier) in short time as compared to the baseline DLN. Also, the linear classifier at every stage is trained only on those instances passed from the previous stage. Since the fraction of input instances passed to the next stage decreases as we go deeper into the network, the training time for the linear classifiers progressively decreases.

*B. Testing the CDLN*

Algorithm 2 describes the overall testing methodology for the CDLN. Given a test instance $I_{test}$, the methodology produces the class label $L_{test}$ for it using $N_{cdl}$. The output from the linear classifier at every stage is monitored to decide if the input can be classified at the current stage or not. For the worst case (very hard instance), all the CNN layers and the corresponding linear classifiers at every stage will be activated and $L_{test}$ will be the class label produced by the final output layer.

---
**Algorithm 2:** Methodology to test the CDLN
---
**Input:** Test instance $I_{test}$, CDLN $N_{cdl}$ with the # of linear classifiers or stages in $N_{cdl}$
**Output:** Class label $L_{test}$
1. Obtain the CNN layer feature vectors for $I_{test}$ ($CNN_i$) corresponding to a stage/layer $i$.
2. If a linear classifier ($LC_i$) is present at stage $i$, obtain the output of $LC_i$ corresponding to $CNN_i$.
3. If the confidence value of the output is beyond a certain threshold δ (user defined), then TERMINATE testing at stage $i$ and Output $L_{test}$ = Class label given by $LC_i$. The layers or stages of $N_{cdl}$ from i+1 onwards are not activated if testing is terminated at stage $i$
4. If the confidence value of the output is below the threshold δ or output has high confidence value for more than one class label, activate the next stage $i+1$.
5. Goto step 1 and repeat this until you reach the final layer of the CDLN.

---

In summary, the design methodology implicitly modulates the number of stages or layers used for classification based on the input and produces an optimal CDLN. The user defined threshold, δ, for the confidence level can be adjusted during runtime to achieve the best tradeoff between accuracy and efficiency improvements of the CDLN. Thus, the proposed approach is systematic and hence can be applied to all image recognition applications.

## IV. EXPERIMENTAL METHODOLOGY

In this section, we describe the experimental setup used to evaluate the performance of the Conditional Deep Learning Network. We have implemented a standard DLN based pattern recognition platform for the MNIST dataset that consists of a 60,000 sample training set and a 10,000 sample test set [17]. We used two different DLN architectures shown in Table I & II as the baseline classifier. The baseline DLNs were trained using the convolutional back-propagation algorithm as proposed in [19]. We employed the training methodology discussed in section III to construct the CDLN (MNIST_2C, MNIST_3C from Table I and II) with optimum number of stages. In addition to the final output layer (FC), MNIST_2C consists of a linear layer of output neurons (O1) after the first pooling layer (P1) and MNIST_3C has output layers (O1, O2) after pooling layers (P1,P2). Please note that the learnt feature vectors from the pooling layers are used as training inputs to the linear classifiers. Addition of linear classifiers enables conditional activation of the layers: (C2, P2, FC) in MNIST_2C and (C2, P2, C3, P3, FC) in MNIST_3C depending upon the difficulty of the input instance.

For hardware execution, we implemented each classifier at the register transfer logic (RTL) level. Synopsys design compiler was used to synthesize the integrated design to a 45nm SOI process from IBM. Finally, Synopsys Power compiler was used to estimate energy consumption of the synthesized netlists.

TABLE I. DLN ARCHITECTURE WITH 6 LAYERS

| BASELINE DLN | SIZE | MNIST_2C (CDL) |
|---|---|---|
| Input (I) | 28x28 | I→C1→P1→C2→ P2→FC ↓ O1 |
| Convolution C1 | 24x24, 6 maps | |
| Pooling P1 | 12x12, 6 maps | |
| Convolution C2 | 8x8, 12 maps | |
| Pooling P2 | 4x4, 12 maps | |
| Fully connected-Output (FC) | 10 | |

TABLE II. DLN ARCHITECTURE WITH 8 LAYERS

| BASELINE DLN | SIZE | MNIST_3C (CDL) |
|---|---|---|
| Input (I) | 28x28 | I→C1→P1→C2→P2→C3→P3→FC ↓ ↓ O1 O2 |
| Convolution C1 | 26x26, 3 maps | |
| Pooling P1 | 13x13, 3 maps | |
| Convolution C2 | 10x10, 6 maps | |
| Pooling P2 | 5x5, 6 maps | |
| Convolution C3 | 3x3, 9 maps | |
| Pooling P3 | 3x3, 9 maps | |
| Fully connected-Output (FC) | 10 | |

## V. RESULTS

In this section, we present the experimental results that establish the benefits associated with CDL.

*A. Energy Improvement*

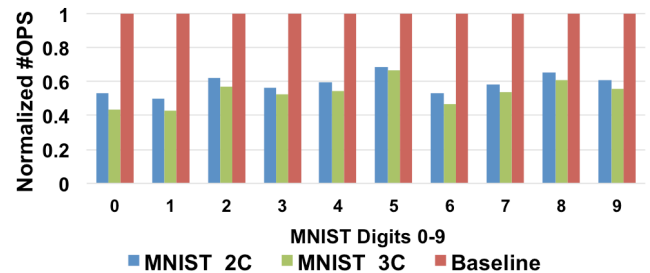

Fig. 5. Normalized OPS for the CDLN (MNIST_2C,MNIST_3C) with respect to baseline

Fig. 5 shows the normalized improvement in efficiency with respect to the DLN (which forms the baseline) for both CDLN (MNIST_2C, MNIST_3C) for all the digits. We quantify efficiency as the average number of operations (or computations) per input (OPS). We observe that while MNIST_2C provides between 1.46x-1.99x (average: 1.73x) improvement in average OPS/input compared to the baseline across the testing set for all digits, MNIST_3C gives 1.50x-2.32x (average: 1.91x) improvement. The higher benefits observed in MNIST_3C can be attributed to the fact that the DLN structure for MNIST_2C is more complex (higher number of neurons and synapses) than MNIST_3C. Additionally, two linear classifiers in MNIST_3C gives the advantage of turning off more layers of the DLN than MNIST_2C for a given input instance. Note that the benefits observed vary for different digits. Fig. 5 clearly illustrates that maximum benefit in both the frameworks is observed for digit 1 and minimum for digit 5. We can thus infer that digit 5 has more hard instances in the testing set that are closer to the non-linear decision boundary and hence need activation of deeper layers for accurate classification. Digit 1, on the other hand, has easier instances away from the non-linear boundary and thus can be classified by the early layers with an approximate linear boundary decision model. In case of hardware implementation, the reduction in OPS translates on an average to 1.84x/1.71x reduction in energy for MNIST_3C and MNIST_2C respectively as shown in Fig. 6.

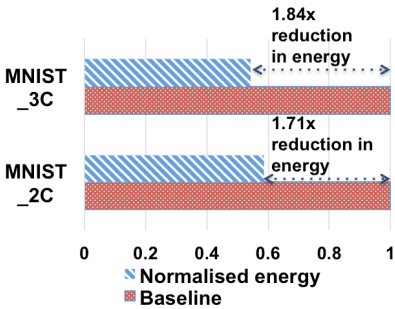 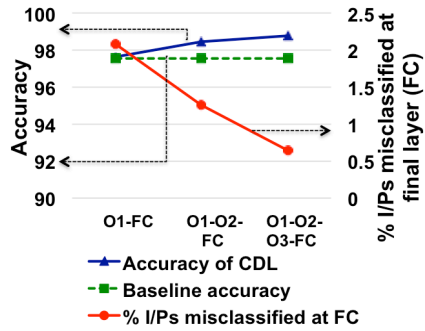 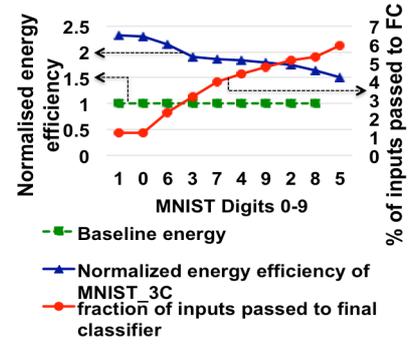

Fig. 6. Normalised energy benefits of CDLN with respect to the baseline

Fig. 7. Accuracy improvement in CDLN with the increase in no. of output layers

Fig. 8. Normalised energy benefits as the difficulty of inputs increases

### B. Enhancement in Accuracy

Table III shows the overall accuracy for the baseline DLN architectures (Table I, II) and the corresponding CDLNs MNIST_2C, MNIST_3C over the testing set. We observe that there is a 1.01% enhancement in accuracy for MNIST_3C compared to baseline and 1.37% for MNIST_2C. The difference in accuracy improvement in both networks is due to the fact that the baseline DLNs in both cases have a different training and test accuracy. In the beginning of the experiment, our motivation behind adding linear classifiers was to get an improvement in efficiency. However, the accuracy enhancement with the CDL methodology implies that the linear classifiers trained by the convolutional layer features perform better than the baseline DLN. We note that the main difference between the baseline DLN and the CDLN is that the baseline DLN uses the last convolutional or pooling layer features as inputs to the fully connected output layer of neurons to predict the classification result. As opposed to using the final fully connected layer for prediction, CDL methodology uses the trained linear classifiers added at every stage or layer which perform their own learning on the input CNN features.

TABLE III. ACCURACY FOR 6-LAYER AND 8-LAYER NETWORKS

| NETWORK | BASELINE | CDLN |
|---|---|---|
| 6- LAYER | 98.04 % | 99.05 % (MNIST_2C) |
| 8-LAYER | 97.55 % | 98.92 % (MNIST_3C) |

As mentioned earlier, the linear classifiers being small can be trained to converge to the global minima (least error attainable by the linear classifier) in a short time as compared to the baseline DLN. It is known that CNN learnt features become more specific and the feature vector also becomes smaller as we go deeper into the network. Thus, it is intuitive that the linear classifiers trained on the latter layers will reach smaller error minima in lesser time than the former ones. Adding more linear classifiers at every layer of the network would progressively minimize the overall error thereby improving the accuracy.

To quantify our theory, we designed an experiment where we added the linear classifiers one at a time with the baseline DLN in Table II. During test time, we monitor the prediction results from the each of the added output layers (O1, O2, O3) in addition to the final output layer (FC) to measure the overall accuracy. Fig. 7 shows the normalized accuracy of the CDLN as we add the output layers one by one at every convolutional layer. We can observe an improvement in accuracy for each of the cases as compared to the baseline (97.55 %). While addition of just one linear classifier (O1-FC) enhances the accuracy by 0.1% (97.65 %), the benefits observed are higher up to 1.4% (98.92 %) with three linear classifiers for every CNN layer of the network. We also observe that the fraction of inputs misclassified by the final layer progressively decreases further corroborating our theory.

### C. Impact of the Difficulty of Inputs on Efficiency

Here, we examine the impact of the variation in the difficulty of input instances on the overall efficiency of the proposed CDLN. Fig. 8 illustrates the normalized energy efficiency observed with MNIST_3C for all the digits in a decreasing order of efficiency. It is evident that the main idea behind CDL is to classify all easy instances in the first stages and enable the latter layers just for the hard instances. From Fig. 8 we can infer that digit 1 and 5 can be broadly categorized as the least and the most difficult instances respectively. Thus, the final output layer (FC) should be enabled more for digit 5 than digit 1. It is clearly seen that FC is activated for only 1% of the total instances of digit 1 and 6% of the more difficult (closer to non-linear decision boundary) digit 5 instances. In the best case, all instances of digit 1 should be classified at the first stage with a linear decision boundary model. However, in practice, there will be some instances that will fall beyond the decision boundary (Fig.1 (b)) and will thus be passed to the next stage. The energy benefits observed decreases as the inputs become more difficult. However, we get an energy benefit of 1.5x compared to the baseline even for the hardest input. We also observe that addition of linear classifiers in MNIST_3C prevents the activation of the deeper layers of the network for nearly 99% of the instances of digit 1. This shows the effectiveness of the CDL methodology.

TABLE IV. IMAGES OF 1 AND 5 CLASSIFIED AT DIFFERENT STAGES

| DIGIT | O1 | O2 | FC |
|---|---|---|---|
| 1 | 1 | 1 | 1 |
| 5 | 5 | 5 | 5 |

Table IV shows some typical examples of the images classified correctly by each output layer (O1, O2, FC) of MNIST_3C for the least and the most difficult digits:-1 and 5 respectively. The table visually supports the fact that easier instances are classified at earlier stages and the difficult ones are passed to the latter stages for correct classification.

### D. Optmizing the Number of Stages in the CDLN

Choosing the right number of stages or linear classifiers added at the CNN layers is critical to the efficiency of the CDL methodology. In the previous section, we observed that the efficiency increases as the fraction of inputs passed to the final output layer (FC) decreases. In order to pass fewer instances to FC, addition of linear classifiers at every CNN layer of the DLN is desirable. Fig. 9 shows the normalized OPS of the CDLN as the output layers are added one at a time for every CNN layer of the DLN architecture from Table II. It is clearly seen that fraction of inputs passed to FC decreases with the increasing number of output stages. We observe a significant drop in the fraction from 42% to 5% with the addition of two output stages (O1-O2-FC). Thus, initially we observe a decrease in #OPS. However, increasing the number of output stages adds an additional overhead to the cost of computation. Note, addition of a third stage (O1-O2-O3-FC) results in a marginal drop in the fraction of inputs passed from 5% to 3%, which is not significant enough to overcome the cost penalty that the addition of the stage imposes. So, we see an increase in #OPS from this point onwards. This break-even point (0.45 #OPS) corresponds to the maximum benefits or lowest #OPS that we can achieve using the CDL methodology for a given baseline DLN. This behavior is taken into account in our design methodology described in the Section III.

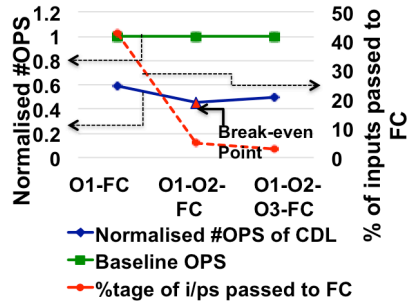

Fig. 9. Normalized #OPS as the no. of stages are increased in CDLN

*E. Efficiency-Accuracy Tradeoff using Confidence Level ($\delta$)*

The linear classifiers in the CDLN, in addition to the class label, provide a class probability. The activation module discussed in Section II compares this probability to the confidence level $\delta$ set by the user to selectively classify the input or pass it to the next stage. Thus, we can regulate $\delta$ to modulate the number of inputs being passed to the latter layers. Fig. 10 shows the variation in the normalized OPS (with respect to baseline DLN which quantifies efficiency) and accuracy for the CDLN (MNIST_3C) with different $\delta$. Setting $\delta$ to a low value implies more input instances will be qualified as hard inputs and passed to the final output layer (FC) for classification. Then, FC will be redundantly activated for the easy inputs as well. As introduced earlier, the accuracy of the CDLN improves as we increase the number of stages. In other words, more inputs should be classified by the linear classifiers at the CNN layers, instead of being passed to FC, for an improvement in accuracy. So, increasing $\delta$ would qualify more inputs as easy instances and result in more inputs being classified at the linear classifiers leading to lesser activation of latter layers. Thus, we see a decrease in #OPS and an increase in accuracy initially. However, beyond a particular $\delta$, a fraction of hard inputs that should be ideally passed to the final layer for classification will now be misclassified at the early stages. This value of $\delta$ (0.5 in Fig. 10) corresponds to the maximum overall accuracy that can be achieved for the given CDLN. Beyond this point, the accuracy would decrease. The # OPS would still continue to decrease with increasing $\delta$. In Fig. 10, we observe that the accuracy increases from 96.12% ($\delta$=0.4) to 99.02% ($\delta$=0.5) while the normalized #OPS reduce from 1.1 to 0.51. Further increase in $\delta$ degrades the accuracy and does not produce a significant reduction in #OPS. Thus, $\delta$ serves as a powerful knob to trade accuracy for efficiency that can be easily adjusted during runtime to get the most optimum results.

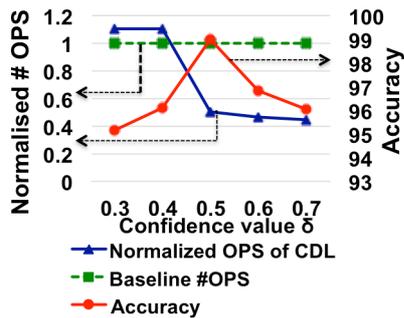

Fig. 10. Efficiency vs. accuracy tradeoff using confidence value $\delta$

## VI. CONCLUSION

Deep learning convolutional neural networks are vital for many computer vision applications and demand significant computational effort in modern computing platforms. In this work, we explore a novel approach to optimize conventional deep learning networks by employing the convolutional layer features to discriminate between easy and hard input data. We propose the concept of Conditional Deep Learning in which easy instances can be classified earlier without activating the latter layers of the network. We achieve this by cascading a linear network of output neurons for each convolutional layer and monitoring the output of the linear network to decide if the classification can be terminated at a current layer or not. The design methodology implictly varies the number of stages or layers used for classification based on the difficulty of the input and produces a CDL with optimal efficiency. To quantify the potential of CDL, we designed the CDLN with two different architectures for the MNIST dataset. Our experiments demonstrate 1.91x reduction in average OPS per input, which translates to 1.84x improvement in energy for the 8-layered DLN (Table II). In addition to energy benefits, our results show that the CDLN yields a better classification accuracy (98.92%) as compared to the corresponding baseline DLN (97.55%).